\pdfoutput=1

\documentclass[11pt]{article}

\usepackage{EMNLP2023}

\usepackage{times}
\usepackage{latexsym}

\usepackage[T1]{fontenc}

\usepackage[utf8]{inputenc}

\usepackage{microtype}

\usepackage{inconsolata}

\usepackage{pifont}
\newcommand{\cmark}{\ding{51}}%
\newcommand{\xmark}{\ding{55}}%
\newcommand{\toolname}{PromptNER }
\usepackage{graphicx}
\usepackage{lscape}

\title{\toolname: Prompting For Named Entity Recognition}

\author{Dhananjay Ashok \\
  Carnegie Mellon University\\
  \texttt{dhananja@cs.cmu.edu} \\\And
  Zachary C. Lipton \\
    Carnegie Mellon University\\
  \texttt{zlipton@cs.cmu.edu} \\}

\begin{document}
\maketitle
\begin{abstract}
In a surprising turn, Large Language Models (LLMs) together with a growing arsenal of prompt-based heuristics
now offer powerful off-the-shelf approaches providing few-shot solutions to myriad classic NLP problems. 
However, despite promising early results, these LLM-based few-shot methods remain far from the state of the art in Named Entity Recognition (NER), where prevailing methods include learning representations via end-to-end structural understanding and fine-tuning on standard labeled corpora. 
In this paper, we introduce \toolname, 
 a new state-of-the-art algorithm for few-Shot and cross-domain NER. To adapt to any new NER task \toolname requires \emph{a set of entity definitions} in addition to the standard few-shot examples. 
Given a sentence, \toolname prompts an LLM to produce  a list of potential entities along with corresponding explanations justifying their compatibility with the provided entity type definitions. 
\toolname achieves state-of-the-art performance on few-shot NER,
achieving a 4\% (absolute) improvement in F1 score on the ConLL dataset, a 9\% (absolute) improvement on the GENIA dataset, and a 4\% (absolute) improvement on the FewNERD dataset.
\toolname also moves the state of the art on Cross Domain NER, outperforming prior methods (including those not limited to the few-shot setting), setting a new mark on $3/5$ CrossNER target domains, with an average F1 gain of $3\%$, despite using less than $2\%$ of the available data.
\end{abstract}

\section{Introduction}
Named Entity Recognition \citep{chinchor1995muc} is often a vital component 
in text processing pipelines for information extraction and semantic understanding
\citep{sharma2022named, ali2022named}. 
Current methods perform well when training data is plentiful
\citep{wang2022deepstruct, yu2020named, li2022unified, wang2020automated}.
However, their applicability to many real-world problems
is hindered by their reliance on fixed entity definitions
and large amounts of in-domain training data
for the specific NER formulation and population of interest.
Unfortunately, commitments about what constitute the relevant entities
vary wildly across use cases, a fact that is reflected
in the diversity of academic datasets
(contrast, e.g., medical NER datasets with CoNLL or OntoNotes). 
Ultimately, these differing commitments 
stem from differences in the envisioned use cases.
Should we categorize the phrase `Theory of General Relativity' as an entity? 
A media company tasked with extracting 
information from political articles
might not designate physical laws
as a relevant class of entities
but a scientific journal might. 
Given the diversity of use cases 
and underlying documents
that characterize different deployment settings,
we might hope ideally for a system to adapt to new settings flexibly,
requiring minimal labeled data, 
human effort, and computational cost.

With the emergence of LLMs, the NLP community 
has developed a repertoire of in-context learning strategies 
that have rapidly advanced the state of few-shot learning
for myriad tasks \citep{brown2020language, wei2022chain, liu2023pre}.
However, such prompting-based approaches have yet 
to show comparable impact in NER, 
where current methods typically cast few-shot learning 
as a domain transfer problem, training on large amounts of source data 
and fine-tuning on exemplars from the target domain \citep{huang2022copner, yang2022factmix}.
Moreover, a significant gap remains between the best few-shot NER methods
and the performance of end-to-end trained models \citep{wang2022deepstruct, xu2022clozing}.
These few-shot methods struggle when the source and target domains 
differ with respect to what constitutes an entity \citep{yang2022factmix, das2022container}. 
A separate class of adaptation methods have shown promise 
when the source and target vary considerably, 
but they tend to require hundreds of data points to be effective
\citep{hu2022label, chen2023one, hu2022entda, chen2022prompt}.


In this paper, we introduce \toolname,
a prompting-based NER method
that achieves state-of-the-art results 
on FewShot NER and CrossDomain NER. 
Our method consists of 4 key components---a 
backbone LLM, a modular definition 
(a document defining the set of entity types), 
a few examples from the target domain, and a precise format for outputing
the extracted entities, which is communicated to the model
via the formatting of the few-shot examples. 
To adapt to a new domain, our method requires modifying
only the definition and the provided examples.
This makes the method flexible and easy to apply across domains.
\toolname achieves 83.48\% F1 score on the CoNLL dataset \citep{sang2003introduction} in a few-shot setting, improving over the best previous few-shot methods by 4\% (absolute). \toolname outperforms the best-competing methods by 9\% (absolute) on the GENIA \citep{kim2003genia} dataset and 4\% (absolute)
on the FewNERD-Intra \citep{ding2021few} setting
and sets a new state of the art on three out of five
of the CrossNER \citep{liu2021crossner} target domains,
despite using only $2\%$ of the available training data.
In ablations, we show that \toolname outperforms
standard Few-Shot Prompting \citep{brown2020language} 
and Chain-of-Thought Prompting \citep{wei2022chain}.

\begin{figure*}[th]
    \centering
    \includegraphics[scale=0.9]{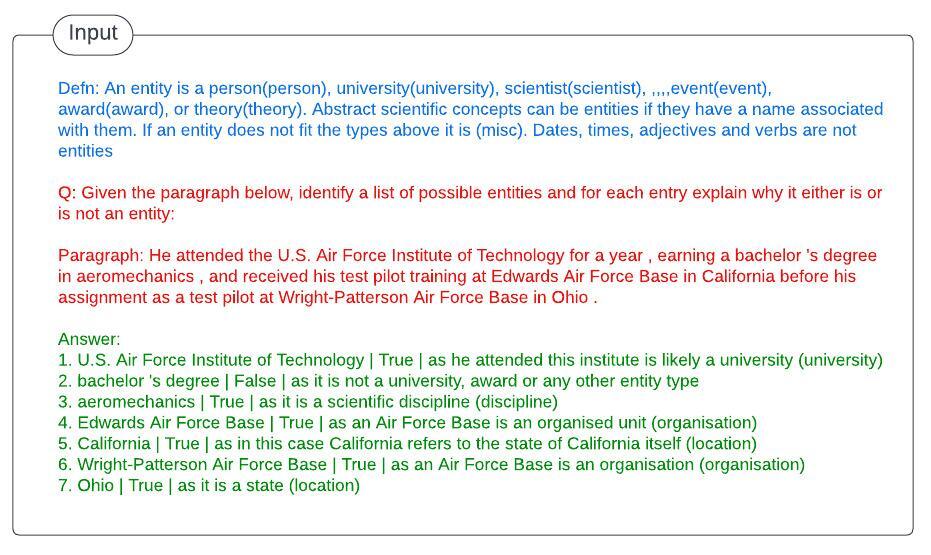}
    \caption{Example of prompt to pre-trained language model. \textcolor{blue}{Definition is in blue}}, \textcolor{red}{question and task in red}, \textcolor{green}{example answer and chain of thought format in green}
    \label{fig:example}
\end{figure*}

\section{Background and Related Works}

\noindent\textbf{Named Entity Recognition} has been well studied
since the formulation of the problem in the mid-90s \citep{chinchor1995muc}, 
with early methods including rule-based systems \citep{eftimov2017rule, farmakiotou2000rule, mikheev1999named}, statistical learning based methods \citep{borthwick1998exploiting, borthwick1999maximum, zhou2002named},
and the use of traditional machine learning methods like SVMs \citep{makino2002tuning, mayfield2003named}. 
With the shift towards deep learning,
RNN-based methods and then transformer-based methods
came to dominate on most NER tasks \citep{de2015survey, huang2015bidirectional, limsopatham2016bidirectional}. 
Most recently, methods leveraging large pre-trained Transformers have advanced the state-of-the-art further \citep{lu2022unified, wang2022deepstruct, yu2020named, tong2022improving, lu2022punifiedner, li2022unified}. 

Recent methods include DeepStruct \citep{wang2022deepstruct},
which modify the typical language modelling procedure 
to make the language model more aware of logical structure in the corpus, 
and then use this trained model for downstream NER. 
Other methods introduce specialized architectures 
with features of the NER problem in mind \citep{yu2020named, li2022unified}.
These methods typically require full dataset access 
and require significant computational resources for training.
Motivated by settings where training on full datasets 
is not possible or practical, some researchers have 
turned their attention to few-shot NER
\citep{church2021emerging, das2022container, huang2022copner}.
Leading few-shot NER methods include approaches which seek to create a pretrained LLM that can then be adapted for NER \citep{wang2022deepstruct, xu2022clozing} and approaches which view NER as a metric learning problem and use prompts to guide the learning process.\citep{huang-etal-2022-copner, chen2022prompt}. The final class of methods we discuss 
tackles the cross domain generalization of NER methods, 
where the training and testing tasks are from different domains, 
these include CP-NER \citep{chen2023one},
which uses collaborative prefix tuning
to learn domain-specific prefixes that can be swapped flexibly to perform NER; 
FactMix \citep{yang2022factmix},
which uses a model-agnostic data augmentation strategy
to improve generalization;
and LANER \citep{hu2022label}, 
which tries to improve transferability of learnt label information.
A few methods try to directly apply in context learning \citep{wei2021finetuned} for NER, 
often interpreting the problem as 
a templated cloze statement infilling problem \citep{lee2021good, cui2021template}.

 \noindent\textbf{Prompting and Chain-of-Thought Prompting} \citep{brown2020language, wei2022chain} 
 represent new ways to use LLMs.
 It has been shown that LLMs can achieve high performance 
 on Few-Shot tasks with a few examples in the context window of the input.
 Chain-of-Thought prompting improves on this approach 
 by providing examples in the prompt 
 which not only contain question answer pairs,
 but also some stated reasoning for the provided answer. 
\textbf{Key Innovations: } Our method, \toolname, 
leverages the power of Chain-of-Thought Prompting
to provide an accurate, flexible, and accessible way 
to perform few-shot NER without requiring 
any parameter updates. 
Our key innovations include the use of
a swappable entity dictionary 
and a new output format.
Together these lead to significant gains 
over the previous state of the art 
in few-shot NER and cross domain NER. 


\begin{table}[t]
\centering
\begin{tabular}{cc}
\hline
Method                      & CoNLL         \\
\hline
COPNER\cite{huang2022copner}                     & $75.8 \pm 2.7$   \\
EntLM\cite{ma2021template}                      & $51.32 \pm 7.67$ \\
FactMix\cite{yang2022factmix}                    & $60.8$          \\
ProML\cite{chen2022prompt}                       & $79.16 \pm 4.49$ \\
UIE\cite{lu2022unified}                        & $67.09$         \\
CONTaiNER\cite{das2022container}                   & $75.8 \pm 2.7$   \\
PMR\cite{xu2022clozing}                        & $65.7 \pm 4.5$   \\
\toolname T5XXL (Us)           &  $45.66 \pm {12.43}$\\
\textbf{\toolname GPT3.5 (Us)}           & $78.62 \pm 4.62$\\
\textbf{\toolname GPT4 (Us)}           & {\bf $\textbf{83.48} \pm \textbf{5.66}$}\\
\hline
\end{tabular}
\caption{FewShot Learning ($0 < k < 5$) on CoNLL dataset. Results show micro-F1 averages and associated standard deviation over 5 runs when available. }
\label{table:conll}
\end{table}

\begin{table*}[th]
\centering
\begin{tabular}{ccccccc}
\hline
Method   & k       & Politics & Literature & Music & AI    & Sciences \\\hline
FactMix\cite{yang2022factmix}  & 100     & 44.66    & 28.89      & 23.75 & 32.09 & 34.13    \\
LANER\cite{hu2022label}    & 100-200 & 74.06    & 71.11      & 78.78 & 65.79 & 71.83    \\
CPNER\cite{chen2023one}    & 100-200 & 76.35    & 72.17      & 80.28 & \textbf{66.39} & \textbf{76.83}    \\
EnTDA\cite{hu2022entda}    & 100     & 72.98    & 68.04      & 76.55 & 62.31 & 72.55    \\
\toolname T5XXL (Us) & \textbf{2}       & 39.43     & 36.55       & 41.93  & 30.67    & 46.32 \\
\textbf{\toolname GPT3.5 (Us)} & \textbf{2}       & 71.74     & 64.15       & 77.78  & 59.35    & 64.83 \\
\textbf{\toolname GPT4 (Us)} & \textbf{2}       & \textbf{78.61}     & \textbf{74.44}       & \textbf{84.26}  & 64.83    & 72.59    \\\hline
\end{tabular}
\caption{Cross Domain results on CrossNER dataset with CoNLL as source domain. $k$ is the number of target domain datapoints used by each method. Results show micro-F1 scores. Despite using only $1\%-2\%$ of the data our method achieves state-of-the-art performance on three of the five datasets}
\label{table:crossner}
\end{table*}

\begin{table}[t]
\centering
\begin{tabular}{cc}
\hline
Method                      & GENIA         \\
\hline
CONTaiNER\cite{das2022container}                   & $44.77 \pm 1.06$   \\
BCL\cite{ming2022few}                     & $46.06 \pm 1.02$   \\
SpanProto\cite{shen-etal-2021-locate}                      & $41.84 \pm 2.66$ \\
PACL                    & $49.58 \pm 1.82$          \\
\toolname T5XXL (Us)           &  $25.13 \pm 3.22$\\
\textbf{\toolname GPT3.5 (Us)}           & {\bf $\textbf{52.80} \pm \textbf{5.15}$}\\
\textbf{\toolname GPT4 (Us)}           & {\bf $\textbf{58.44} \pm \textbf{6.82}$}\\
\hline
\end{tabular}
\caption{FewShot Learning ($0 < k < 5$) on GENIA dataset. Results show micro-F1 averages and associated standard deviation over 5 runs when available. }
\label{table:genia}
\end{table}

\section{Method}
In NER, we are given a sentence 
and asked to predict the set of entities that it contains. 
We are sometimes given a natural language description of what we would like to consider an entity, or a list of possible entity types.

Our method consists of 3 key components ~\ref{fig:example}:

\textbf{Conditional Generation Language Model}:
we leverage the power of pretrained LLMs 
which have been trained on conditional generation tasks. 
This is a departure from several works in the field 
that consider the NER problem as a sequence to label (or discriminative) problem 
\citep{ali2022named, sharma2022named}, 
or the prompting-based approaches
that consider NER to be a cloze statement prediction problem \citep{cui2021template, lee2021good}.
We believe that a sequence-to-sequence formulation of NER 
has an advantage over these other formulations
because it resolves the combinatorial explosion 
which occurs when entities can be more than a single word long (e.g "United Nations"). 
Discriminative and in-filling approaches would have to iterate 
over all possible combinations of words,
which includes either uncomfortable manual thresholding of the maximum token length of an entity,
or requires a search over all n-grams over the sentence. 
Additionally, using general Seq-to-Seq models 
opens up modeling possibilities, allowing us 
to pursue strategies based on chain-of-thought-like reasoning.
These possibilities are not readily available in the more traditional
aligned sequence tagging architectures traditionally used for NER.

\textbf{Modular Definitions:} the implicit commitments
about what constitute instances of the relevant entity types
can vary wildly across different NER settings.
The difficulty of capturing these subtleties in just a few examples
can hinder typical few-shot approaches to NER.
In our approach to few-shot, NER, each problem is defined
not only by a small set of exemplars (the few-shot examples)
but also by a per-domain definition. 
Here, the modular definition consists of 
a natural language description of what does (and does not) constitute an entity.
This can be useful in instances where the typical natural language connotation
of the word `entity' may include concepts the specific NER task would want to exclude.
Since this document is composed in natural language,
it can easily be composed by an end user with no technical knowledge.

\textbf{Potential Entity Output Template:} recent work on Chain-of-Thought Prompting has shown that incentivizing the model to output its answer in a structure which facilitates reasoning like steps along with the final answer can significantly boost the performance of large language models \citep{wang2022towards, wei2022chain}. Motivated by this, we create a template structure for the output of the LLM which allows it to emulate reasoning to decide whether a given phrase is classified as an entity or not as well as what type of entity it is. The exact structure is one where each line of the output mentions a distinct candidate entity, a decision on whether or not the candidate should be considered an entity and an explanation for why or why not along with what entity type it belongs to. In the ablation section, we show that the inclusion of examples and a clear definition are the most important parts of this NER system. 

Seen as a whole, this pipeline provides flexibility with little cost. The only components that need to change to adapt the method to new domains are the definition and examples, both of which have virtually no computational cost and do not require an entire data collection initiative to gather. There are many situations where such a flexible solution is preferred over methods that require fine-tuning of parameters \citep{ding2023parameter, wei2021finetuned}, Since the method makes no assumptions on the architecture or specifics of the backbone LLM, it can quickly be upgraded if a new generation of more powerful LLMs were to be released \citep{liu2023pre, kaplan2020scaling}.

In the following sections we demonstrate the predictive power of this method across a variety of settings. 
 \footnote{Code available at \url{https://anonymous.4open.science/r/PromptNER-2BB0/ }}

\begin{table}[th]
\centering
\begin{tabular}{ccc}
\hline
Method        & k & FewNERD                        \\\hline
ProML\cite{chen2022prompt}         & 5 & 68.1 $\pm$ 0.35  \\
CONTaiNER\cite{das2022container}     & 5 & 47.51                          \\
Meta Learning\cite{ma2022decomposed} & 5 & 56.8 $\pm$  0.14 \\
\toolname T5XXL (Us) & \textbf{2} & 55.7 $\pm$  1.09 \\
\textbf{\toolname GPT3.5 (Us)}      & \textbf{2} & ${62.33} \pm {6.30}$\\
\textbf{\toolname GPT 4 (Us)}      & \textbf{2} & $\textbf{72.63} \pm \textbf{5.48}$ \\\hline
\end{tabular}
\caption{Few Shot results on the FewNERD dataset on the INTRA 10-way task, $k$ is the number of datapoints used by each method. Results show micro-F1 scores. Our method manages to outperform other methods by $4\%$ on F1-score}
\label{table:fewnerd}
\end{table}

\section{Experiments and Results}

\noindent\textbf{Resource and Model Description}

For all the experiments below we present the result of our method when using a Pretrained Conditional Generation Language Model of T5-Flan (11B) \citep{chung2022scaling}, GPT-3.5 
 \citep{brown2020language} (specifically the text-davinci-003 model) and GPT4 \citep{openai2023gpt4}. The results for all competing methods are taken from the tables reported in their respective publications and papers. We use the standard metric in NER \citep{de2015survey, wang2022deepstruct} Micro-F1 score and report the mean and variance over 5 runs on a random sample of 500 examples on the test set.

\noindent\textbf{Experiment 1: Standard Low Resource NER}

In this experiment we used the most common NER dataset: CoNLL \citep{sang2003introduction}. This dataset is one where the standard methods are capable of reaching F1-score performance in the range of $91\%-94\%$, \citep{wang2022deepstruct} however, when performed in the low resource settings, these methods are significantly less powerful. We show results for Few Shot Learning with $k=5$ points from CoNLL for all the competing methods in table~\ref{table:conll} and report the standard averaged micro-F1 scores of the models. The table shows that our method outperforms all competing methods when evaluated in the low resource regime - with GPT4 achieving absolute performance gains of around $4\%$ in F1 score. 

\noindent\textbf{Experiment 2: Cross Domain NER} 

In this experiment we use the CrossNER \citep{liu2021crossner} dataset. The training set is a subset of the CoNLL dataset \citep{sang2003introduction}, but the evaluation domains are from five domains with different entity types - Politics, Literature, Music, AI, and Natural Sciences. The dataset does not have any explicitly contradictory examples across splits (i.e. there is no phrase that is considered an entity in one domain that is explicitly marked as not an entity in another domain), however often the conception of what is an entity can vary significantly. For example,  in the AI split abstract mathematical algorithms and methods in Machine Learning like the phrase `deep learning' are considered entities, while in Politics abstract political ideologies and methods in Political Science like `polling' are not entities. On this dataset, few shot methods are not typically able to perform well with only around 5-10 examples, and hence all other successful methods use a significant portion of the training and dev splits of CrossNER (100-200 examples). We show our results in table~\ref{table:crossner}, and despite using only $1\%-2\%$ of the data as the other methods, we are able to achieve state-of-the-art performance in $3/5$ of the domains with GPT4, outperforming other methods by an absolute F1 of $3\%$ on average.The systems seem to perform worse on the AI and Science domain, however it is difficult to make any inferences on the reason why this occurs. 

\noindent{\textbf{Experiment 3: Biomedical Domain NER}} 

We next use the GENIA dataset \citep{kim2003genia}, a biomedical dataset taken from medical database. This presents a different domain with a significant shift in vocabulary in the corpus when compared to the previous datasets. This dataset also has more entity types (32 vs 17) and is more technical than the CrossNER Natural Sciences domain. We show our results on the 32-way 5 shot setting in table~\ref{table:genia}. Once again \toolname outperforms all competing methods, with GPT3.5 outperforming the best competitor by 3\% absolute F1 and GPT4 doing so by 9\%. This shows our method is flexible, and can perform well across a variety of domains with little overhead.

\noindent\textbf{Experiment 4: Contradictory Domain NER} 

In this experiment we use the Intra split of the FewNERD dataset \citep{ding2021few}, using the test split and compiling results for the 10-way problem setting. This dataset is contradictory, in that the sets  marked train, dev and test all have non-overlapping entity types which are labeled as entities. For example the train split considers people to be entities, but does not consider events or buildings to be entities, while the dev set considers only buildings or events to be entities. This is a difficult benchmark as the labels of the training and dev sets are actively misleading when the goal is to perform well on the test set. Table ~\ref{table:fewnerd} shows how our method outperforms all competing methods in this setting, doing so by an average (absolute) percentage increase of over $4\%$ F1. This shows the flexibility of our method in even the most pathological of cases: where the requirement for entities to be extracted is actively in conflict with the common understanding of the word `entity', and the existing data sources. Changing the definition to explicitly rule out objects which would normally be considered an entity is particularly useful here, as shown in later ablations.

\noindent{\textbf{Data Contamination Concerns: }}
Since we are using Large Language Models which have been trained on data from undisclosed sources \citep{brown2020language, openai2023gpt4}, we consider the possibility that parts of the evaluation sets from the above experiments has been seen by the model during the language modeling period, and the implications of this on the results of the experiments. Recent results \citep{chowdhery2022palm} suggest that when using Large Language Models for Few-Shot Learning, the performance on contaminated data (which has been seen during training) is not significantly different from the performance on clean, unseen data. 

The underlying text corpora for all these datasets are sourced from easily accessible collections of corpora online (e.g. Reuters articles) and so is quite likely to have been seen during the training period. This is not too problematic for the NER task in specific, as seeing the sentences with a language modeling objective does not by itself provide any signal on which words are named entities, hence we are more concerned with the possibility of the label information being seen during language modeling. Since we cannot reliably say that there was not some explicit entity-aware training segment during the GPT3 to GPT4 series, we rely on the published dates on which GPT3.5 stops updating its information. The ConLL and GENIA datasets have been available for over 10 years and have been extremely popular datasets for a long time, implying that the label set may have been seen during Language Modelling. There is still valuable insights to obtain from the experiment, however, as GPT4 has significantly better performance than GPT3.5 on both datasets, implying that even if both methods have seen the label information, there are benefits to scaling and using a more powerful backbone Language Model. CrossNER was released in Dec 2020, and was published as \citet{liu2021crossner} only 6 months before the official GPT3.5 knowledge cutoff date. It is also a much less popular NER dataset, being used only by methods seeking to show domain transferability of their systems, as such we argue it is unlikely that explicit label information from this dataset was used during GPT3.5 training. Finally FewNERD was released only a few months before the official cut-off date for GPT3.5 data of September 2021. This makes it highly unlikely that FewNERD label data was seen in the training process of GPT3.5. Overall we cannot make any strong claims about data contamination, however we can say that for the CrossNER and especially the FewNERD experiments, it is unlikely that data contamination is the primary driving force behind the improvement in the results on these datasets. 

\begin{table*}[ht]
\begin{tabular}{|l|l|l|l|l|l|l|l|l|l|}
\hline
Model & Model Size  & ConLL                              & Genia                              & Politics                           & Literature                         & Music                              & AI                                 & Science                            & FewNERD                            \\\hline
GPT4  & \textbf{?}  & \textbf{83.48} & \textbf{58.44} & \textbf{78.61} & \textbf{74.44} & \textbf{84.26} & \textbf{64.83} & \textbf{72.59} & \textbf{72.63} \\
GPT3  & 175 Billion & 78.62                              & 52.8                               & 71.74                              & 64.15                              & 77.78                              & 59.35                              & 64.83                              & 62.33                              \\
T5XXL & 11 Billion  & 45.66                              & 19.34                              & 39.43                              & 36.55                              & 41.93                              & 30.67                              & 46.32                              & 23.2                               \\
T5XL  & 3 Billion   & 24.12                          & 10.5                               & 18.45                              & 18.62                              & 25.79                 & 10.58                              & 26.39                              & 8.35                                \\\hline
\end{tabular}
\caption{Model performance over various model sizes, there are clear benefits to scaling the backbone Large Language Model}
\label{table:size}
\end{table*}

\begin{table*}[ht]
\centering
\begin{tabular}{|l|l|l|l|l|l|l|l|l|l|l|l||l|}
\hline
Def & FS & CoT   & Cand & ConLL         & Genia         & Pol      & Lit    & Mus         & AI            & Sci       & FewNERD       & Avg Rank \\\hline
\textcolor{teal}{\cmark}        & \textcolor{teal}{\cmark}     & \textcolor{teal}{\cmark}  & \textcolor{teal}{\cmark}       & \textcolor{teal}{\textbf{78.6}} & \textcolor{teal}{\textbf{52.8}} & \textcolor{teal}{\textbf{71.7}} & \textcolor{teal}{\textbf{64.1}} & \textcolor{teal}{\textbf{77.7}} & \textcolor{teal}{\textbf{59.3}} & \textcolor{teal}{\textbf{64.8}} & \textcolor{teal}{\textbf{62.3}} & \textcolor{teal}{1}                            \\
\cmark        & \cmark     & \cmark  & \xmark      & 71.6          & 38.5          & 61.3          & 46.3          & 60.2          & 34.2          & 46.8          & 57.3          & 3.5                          \\
\cmark        & \cmark     & \xmark & \cmark       & 75.1          & 49.2          & 70.4          & 54.9          & 70.6          & 53.6          & 60.5          & 42.4          & 2.1                          \\
\cmark        & \xmark    & \cmark  & \cmark       & 68.1          & 23.2          & 20.3          & 21.3          & 24.5          & 40.7          & 40.6          & 34.6          & 5.6                          \\
\xmark       & \cmark     & \cmark  & \cmark       & 63.3          & 46.2          & 57.7          & 49.6          & 50            & 29            & 50.8          & 34.8          & 4                            \\
\textcolor{blue}{\xmark}       & \textcolor{blue}{\cmark}     & \textcolor{blue}{\cmark}  & \textcolor{blue}{\xmark}      & \textcolor{blue}{54.8}          & \textcolor{blue}{37.2}          & \textcolor{blue}{49.8}          & \textcolor{blue}{37.3}          & \textcolor{blue}{54.7}          & \textcolor{blue}{27.8}          & \textcolor{blue}{21.7}          & \textcolor{blue}{18.8}          & \textcolor{blue}{5.6}                          \\
\textcolor{red}{\xmark}       & \textcolor{red}{\cmark}     & \textcolor{red}{\xmark} & \textcolor{red}{\xmark}      & \textcolor{red}{49.7}          & \textcolor{red}{39.3}          & \textcolor{red}{42.5}          & \textcolor{red}{40.3}          & \textcolor{red}{48.6}          & \textcolor{red}{24.5}          & \textcolor{red}{35.9}          & \textcolor{red}{16.1}          & \textcolor{red}{6.1}                         
 \\\hline
\end{tabular}
\caption{Ablation over components of \toolname on GPT3.5. Def: Definitions, FS: Few Shot Examples, CoT: Explanations required, Cand: Candidate entities in predicted list. Every component improves performance of the method in general, with the setting of \textcolor{teal}{all components} vastly outperforming the traditional \textcolor{red}{Few Shot Prompting} and \textcolor{blue}{Chain-of-Thought Prompting} methods}
\label{table:gpt3_ablation}
\end{table*}

\section{Ablations and Variations}
In this section, we set out to investigate the effects of the different aspects of this pipeline and answer some questions on what really matters to the success of the method: In all the tables for this section we refer to the CrossNER Domains by their domain name alone and we refer to the FewNERD Test Intra 10 way setting as FewNERD alone. 

\noindent\textbf{Pretrained Language Model:} We hold all other parts of the pipeline constant i.e the definitions, examples and chain of thought structure of the output and vary the baseline model that we use to compute our predictions, we hope to understand whether there is any trend over which models perform better. Table ~\ref{table:size} shows the results that we might expect, there are significant gains to scaling the size of the Large Language Model. A qualitative analysis of the results suggests that T5XL is barely able to perform the instruction provided in the prompt, sometimes breaking the output format structure, often predicting a single entity multiple times, etc. T5XXL is much better, it is able to follow the output format consistently, however, it is unable to use the definition of entities properly, often labeling dates, numbers, and months as entities despite being explicitly provided a definition that excludes these types of words. This gives us reason to believe that this method is likely to improve as LLMs get better at following instructions more exactly.

\noindent\textbf{Components of \toolname:} We can consider \toolname as having 4 different components that can sequentially be turned off - the provision of a definition (Defn), the provision of few-shot examples (FS), the requirement to explain the reasoning on why a candidate is or is not an entity (CoT) and finally whether or not the list should contain only entities, or also contain candidate entities that may be declared as not meeting the definition of an entity (Cand). In tables~\ref{table:gpt3_ablation} we show how the performance changes for GPT3.5 as we remove only one component of the system and specifically check configurations of Chain-of-thought Prompting and FewShot Prompting. For the sake of brevity, we only show the mean of the Micro-F1, however, a complete table with standard deviations can be found in the appendix~\ref{table:full_ablation}. The results consistently show that every part of the pipeline is useful, with the definition being more important in tasks where the common conception of an entity is quite different from that of the domain (FewNERD, AI) as opposed to the more standard setting (ConLL, Politics). Notably, the row that corresponds to classic Few Shot Learning has an average rank of $6.1$, with classic Chain of Thought Prompting having an average rank of $5.6$, both of these are greatly outperformed by the setting with all components included. This shows that the inclusion of definitions and candidates offers benefits over classical prompting-based approaches. 
The results for GPT4~\ref{table:gpt4_ablation} show very similar trends, with again the row corresponding to all components being included having a much better average rank than all other settings. We also see that of all components of the system, the removal of the definitions and the examples are the most damaging across all the datasets. 

\begin{table*}[ht]
    \centering
    \begin{tabular}{|l|l|l|l|l|l|l|l|l|l|l|l||l|}
    \hline
    Def & FS & CoT   & Cand & ConLL         & Genia         & Pol      & Lit    & Mus         & AI            & Sci       & FewNERD       & Avg Rank \\\hline
\textcolor{teal}{\cmark}        & \textcolor{teal}{\cmark}     & \textcolor{teal}{\cmark}  & \textcolor{teal}{\cmark}       & \textcolor{teal}{83.4}          & \textcolor{teal}{\textbf{58.4}} & \textcolor{teal}{\textbf{78.6}} & \textcolor{teal}{\textbf{74.4}} & \textcolor{teal}{\textbf{84.2}} & \textcolor{teal}{64.8}          & \textcolor{teal}{\textbf{72.5}} & \textcolor{teal}{\textbf{72.6}} & \textcolor{teal}{1.2}      \\
\cmark        & \cmark     & \cmark  & \xmark      & 78.5          & 51.8          & 70.3          & 69.6          & 73.8          & \textbf{66.1} & 67.6          & 59.5          & 2.5      \\
\cmark        & \cmark     & \xmark & \cmark       & 67.4          & 49.2          & 72.4          & 63.5          & 80.5          & 60.5          & 59.7          & 62.8          & 3        \\
\cmark        & \xmark    & \cmark  & \cmark       & \textbf{84.3} & 30.2          & 62.8          & 53.2          & 63.5          & 42.7          & 41.4          & 30.2          & 4.7      \\
\xmark       & \cmark     & \cmark  & \cmark       & 70.7          & 53.3          & 65.6          & 55            & 53            & 45.1          & 53.7          & 43.2          & 4.1      \\
\textcolor{blue}{\xmark}       & \textcolor{blue}{\cmark}     & \textcolor{blue}{\cmark}  & \textcolor{blue}{\xmark}      & \textcolor{blue}{63}            & \textcolor{blue}{38.5}          & \textcolor{blue}{60.2}          & \textcolor{blue}{54.6}          & \textcolor{blue}{58.9}          & \textcolor{blue}{35.9}          & \textcolor{blue}{46.3}          & \textcolor{blue}{27.1}          & \textcolor{blue}{5.6}      \\
\textcolor{red}{\xmark}       & \textcolor{red}{\cmark}     & \textcolor{red}{\xmark} & \textcolor{red}{\xmark}      & \textcolor{red}{66.7}          & \textcolor{red}{27.4}          & \textcolor{red}{58.3}          & \textcolor{red}{46.4}          & \textcolor{red}{54.7}          & \textcolor{red}{27.7}          & \textcolor{red}{36.2}          & \textcolor{red}{21.9}          & \textcolor{red}{6.2}  
 \\\hline
\end{tabular}
\caption{Ablation over components of CoTNER on GPT4. Def: Definitions, FS: Few Shot Examples, CoT: Explanations required, Cand: Candidate entities in predicted list. Every component improves performance of the method in general, with the setting of \textcolor{teal}{all components} vastly outperforming traditional \textcolor{red}{Few Shot Prompting} and \textcolor{blue}{Chain-of-Thought Prompting}. Candidate inclusion is the single most important component}
\label{table:gpt4_ablation}
\end{table*}

\begin{table}[]
\begin{tabular}{ccccccc}\hline
Survey                  & All & P & L & M & A & S \\\hline
\toolname Acc                 & 44  & 60  & 30  & 65  & 35 & 30  \\
Labels Acc                  & 52  & 35  & 80  & 65  & 40 & 40  \\
\toolname better   & 33  & 60  & 20  & 20  & 30 & 35  \\
\toolname worse & 56  & 15  & 60  & 60  & 65 & 50 \\\hline
\end{tabular}
\caption{Survey Results with the CrossNER datasets as columns, all numbers are percentages. \toolname Acc and Labels Acc show the fraction of these lists marked as having a correct answer. Better and worse rows show the percentage of times annotators selected \toolname as the better or worse list. Results show there are many cases where the `incorrect predictions' of \toolname should be considered as good as the ground truth. }
\label{table:survey}
\end{table}

\section{Human Survey of Errors}
An inspection of the errors of \toolname shows many cases where the ground truth and the prediction are in fact different, however, it could be argued that the difference is either meaningless or that given the under-specified nature of NER both sets of entities are equally valid. To quantify the extent to which this is true we randomly picked 20 examples from each CrossNER dataset where the prediction and ground truth entity lists differ (the set of entities identified in a sentence, ignoring type) and asked 10 human annotators (each example is seen by 3 distinct annotators) to comment on the lists. Specifically, the annotators are given a dataset dependent definition of the NER problem and are shown two lists of entities. One of these lists is the set of entities identified by our GPT4 method, the other is the ground truth, however, evaluators are not given this information. They identify if the lists have correctly identified all the named entities and only named entities in a sentence, and if not are asked to provide the phrases that are incorrectly identified. Finally, they are asked for an opinion on which list (if any) can be considered better at identifying which words or phrases are named entities. 

The results of the survey are summarized in Table~\ref{table:survey}. In all of the domains, at least 20\% of the examples are ones where \toolname has the better entity list and \toolname is worse than the ground truth label no more than 40\% of the time. In 4/5 of the domains, the difference between the accuracy of the ground truth labels and \toolname does not exceed 10\%. There is interestingly a domain (Politics) where \toolname is considerably better than the ground truth labels. This survey confirms that there are many legitimate errors in the \toolname predictions, with the ground truth label consistently providing a better option. The survey also shows, however,  that even before we consider the entity type annotations, a very considerable percentage of the disagreeing predictions made by \toolname are not `mistakes' but rather equivalent and roughly equally acceptable solutions to the NER instance. The results suggest that the F1 metric scores reported in Table~\ref{table:crossner} are estimates with a downward bias, especially in domains like Politics and Sciences, where the annotators were not likely to say that the \toolname prediction was worse than the ground truth labels. More generally the survey exposes a need for human evaluation and a measure of inter-annotator disagreement on NER datasets, as on datasets with high disagreement it may be less fruitful to expect models to approach the 100\% F1 mark. 

\section{Limitations}
Due to its reliance on prompting an LLM, \toolname is unable to preserve span information and represent the spans of the words or phrases in the LLM output. This then necessitates a hand-crafted parsing strategy to extract the predicted entity spans and types from the LLM output, which leads to suboptimal solutions when there are many repeated words or phrases in a sentence. The experiments in this paper are not able to explicitly control for data contamination, unfortunately, this is unavoidable when using pre-trained LMs from sources that do not reveal the data they trained on. There is also a danger of interpreting parts of this pipeline incorrectly for the sake of interpretability --- the explanations provided by the system can be logically inconsistent and should not be considered a way to make the NER system interpretable, the candidate lists can completely miss entities from a sentence and so should not be considered as a safe shortlist which will contain all entities with high probability. 

\section{Conclusion}
In this paper, we introduce \toolname, an NER system that outperforms competing few-shot methods on the ConLL, GENIA and FewNERD datasets, outperforms cross-domain methods on 3/5 splits of CrossNER, and has superior performance than few-shot and chain-of-thought prompting on GPT4. We also conduct a human study of the disagreements between the ground truth of the CrossNER dataset and our model output, finding that a notable percentage of the disagreements are not ones that are considered `mistakes' by a human annotator. Overall we provide an alternate way to approach the few-shot NER problem that requires no specialized pretraining, is easy to adjust to different domains, and maintains performance despite using very little training data. 

\bibliography{anthology,custom}
\bibliographystyle{acl_natbib}

\newpage 

\newpage

\appendix

\section{Appendix}
\label{sec:appendix}
\begin{landscape}
\begin{table}[]
\begin{tabular}{cccccccccccc}
\hline
Definitions & Few Shot & CoT   & Candidates & ConLL                  & Genia            & Politics               & Literature            & Music                    & AI                    & Science & FewNERD                \\\hline
\cmark        & \cmark     & \cmark  & \cmark       & 83.4 $\pm$ 5.6          & \textbf{58.4 } $\pm$ \textbf{ 6.8} & \textbf{78.6 } $\pm$ \textbf{ 4.7} & \textbf{74.4 } $\pm$ \textbf{ 6.2} & \textbf{84.2 } $\pm$ \textbf{ 5.1} & 64.8 $\pm$ 6.8          & \textbf{72.5 } $\pm$ \textbf{ 4.1} & \textbf{72.6 } $\pm$ \textbf{ 5.4} \\
\cmark        & \cmark     & \cmark  & \xmark      & 78.5 $\pm$ 5.2          & 51.8 $\pm$ 5.2          & 70.3 $\pm$ 3.1          & 69.6 $\pm$ 4.1          & 73.8 $\pm$ 2.5          & \textbf{66.1 } $\pm$ \textbf{ 4.6} & 67.6 $\pm$ 4.1          & 59.5 $\pm$ 3.6          \\
\cmark        & \cmark     & \xmark & \cmark       & 67.4 $\pm$ 3.4          & 49.2 $\pm$ 3.6          & 72.4 $\pm$ 5.1          & 63.5 $\pm$ 5.6          & 80.5 $\pm$ 6.4          & 60.5 $\pm$ 4.7          & 59.7 $\pm$ 3.5          & 62.8 $\pm$ 4.7          \\
\cmark        & \xmark    & \cmark  & \cmark       & \textbf{84.3 $\pm$ 2.4} & 30.2 $\pm$ 7.9          & 62.8 $\pm$ 5.2          & 53.2 $\pm$ 6.7          & 63.5 $\pm$ 5.6          & 42.7 $\pm$ 3.8          & 30.2 $\pm$ 7.9          & 30.2 $\pm$ 7.9          \\
\xmark       & \cmark     & \cmark  & \cmark       & 70.7 $\pm$ 5.8          & 53.3 $\pm$ 6.1          & 65.6 $\pm$ 4.3          & 55 $\pm$ 2.35           & 53 $\pm$ 5.8            & 45.1 $\pm$ 4.2          & 53.3 $\pm$ 6.1          & 43.2 $\pm$ 6.3          \\
\xmark       & \cmark     & \cmark  & \xmark      & 63 $\pm$ 4.72           & 38.5 $\pm$ 6.2          & 60.2 $\pm$ 6.1          & 54.6 $\pm$ 5.7          & 58.9 $\pm$ 5.3          & 35.9 $\pm$ 4.6          & 46.3 $\pm$ 7.5          & 27.1 $\pm$ 5.3          \\
\xmark       & \cmark     & \xmark & \xmark      & 66.7 $\pm$ 7.0          & 27.4 $\pm$ 5.2          & 58.3 $\pm$ 5.2          & 46.4 $\pm$ 5.1          & 54.7 $\pm$ 2.4          & 27.7 $\pm$ 6.3          & 36.2 $\pm$ 5.3          & 21.9 $\pm$ 4.5          \\\hline
\end{tabular}
\caption{Complete Ablation Results for GPT4}
\label{table:full_ablation}
\end{table}
\end{landscape}

\subsection{Exact Prompts used for each dataset}

\begin{figure*}[th]
    \centering \includegraphics{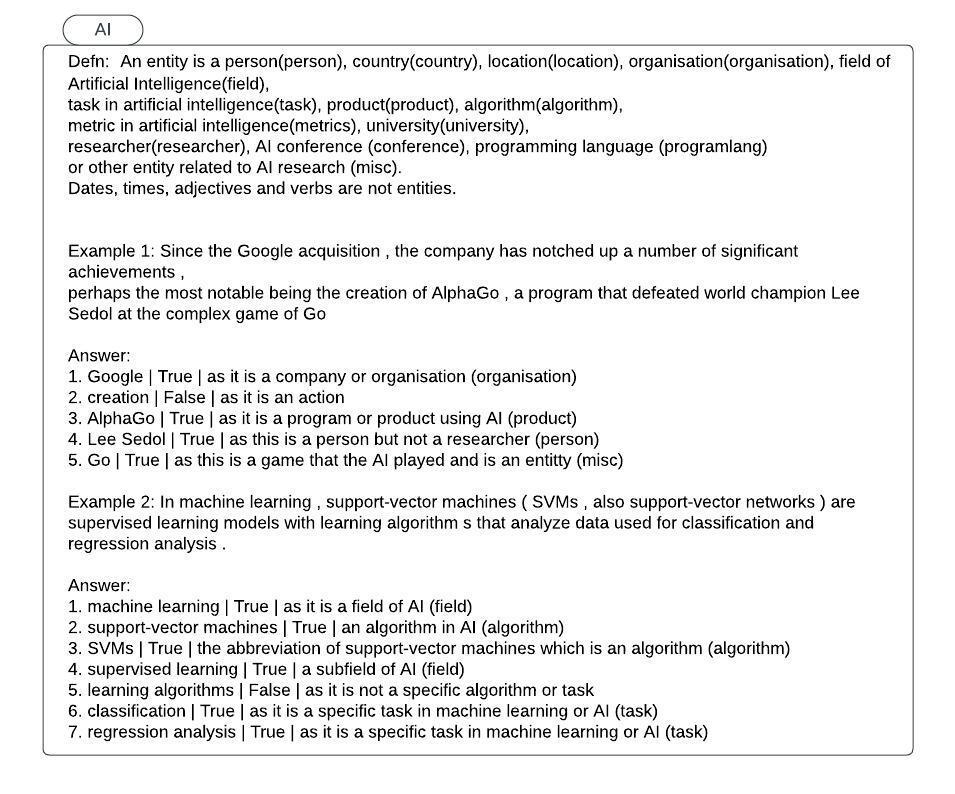}
    \caption{AI Input Prompt}    \label{fig:input_prompts}
\end{figure*}

\begin{figure*}
    \centering \includegraphics{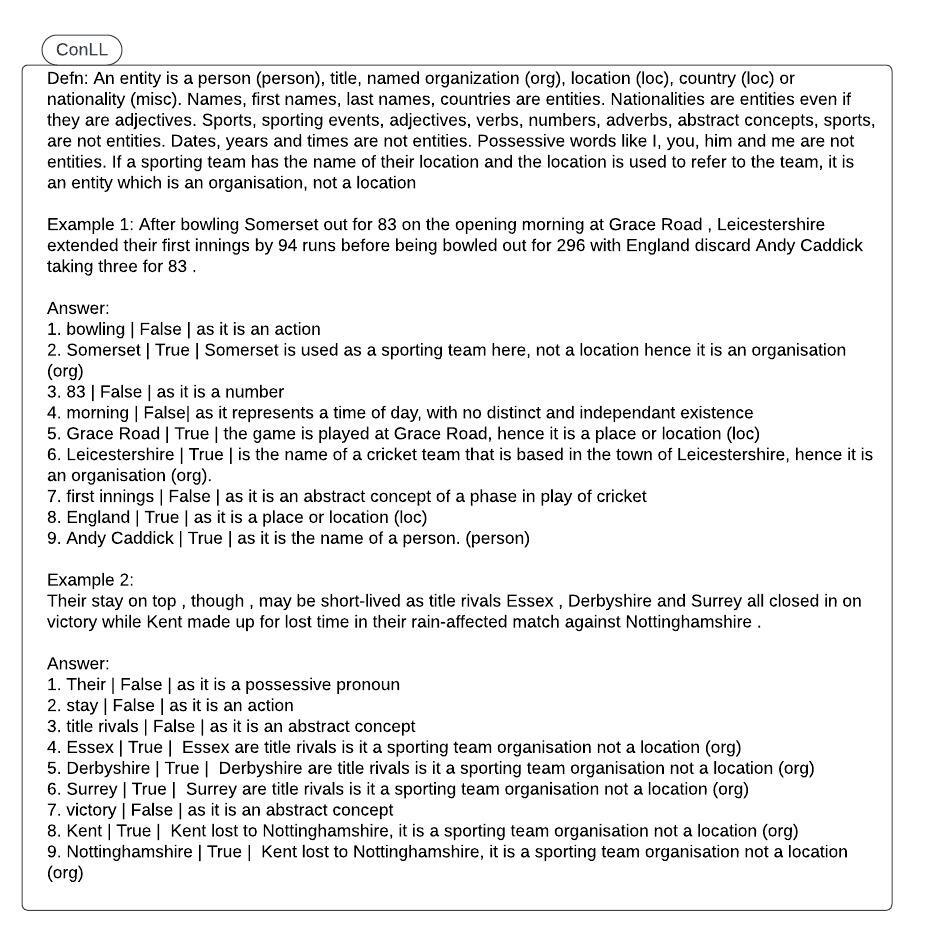}
    \caption{Conll Input Prompt}
\end{figure*}

\begin{figure*}
    \centering \includegraphics{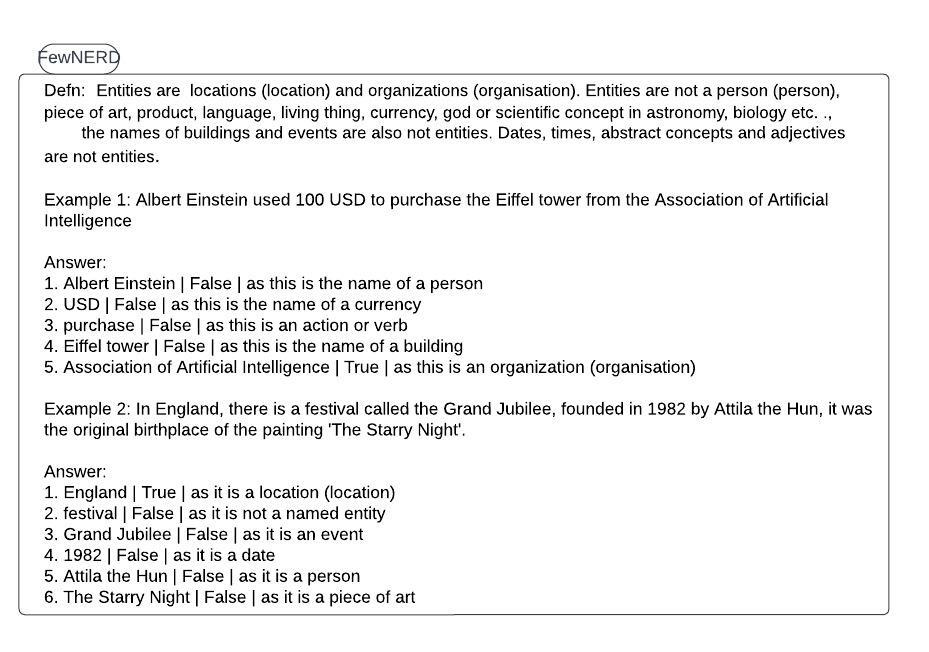}
    \caption{FewNERD Input Prompt}
\end{figure*}

\begin{figure*}
    \centering \includegraphics{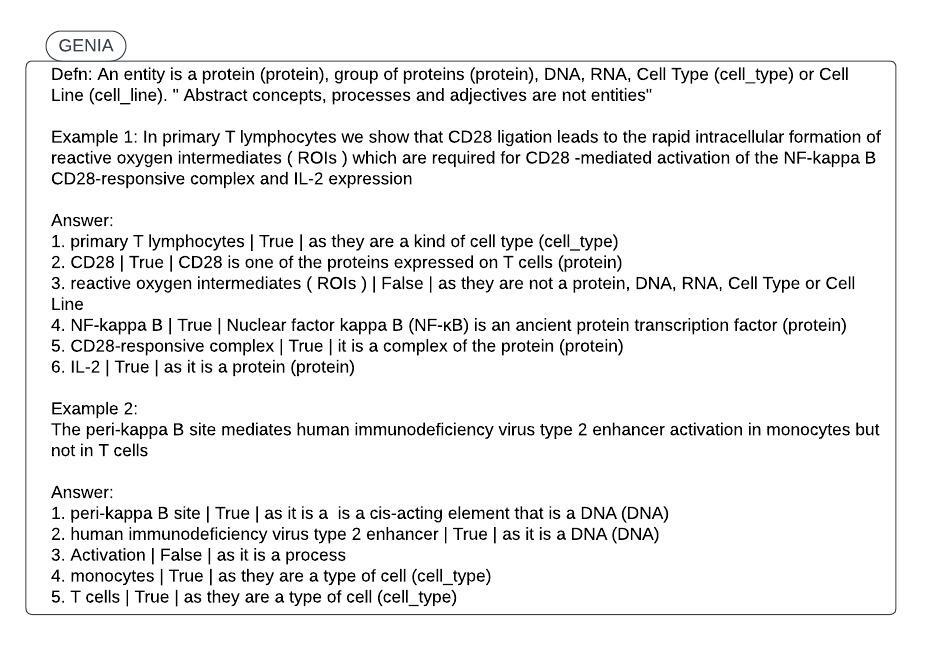}
    \caption{GENIA Input Prompt}
\end{figure*}

\begin{figure*}
    \centering \includegraphics{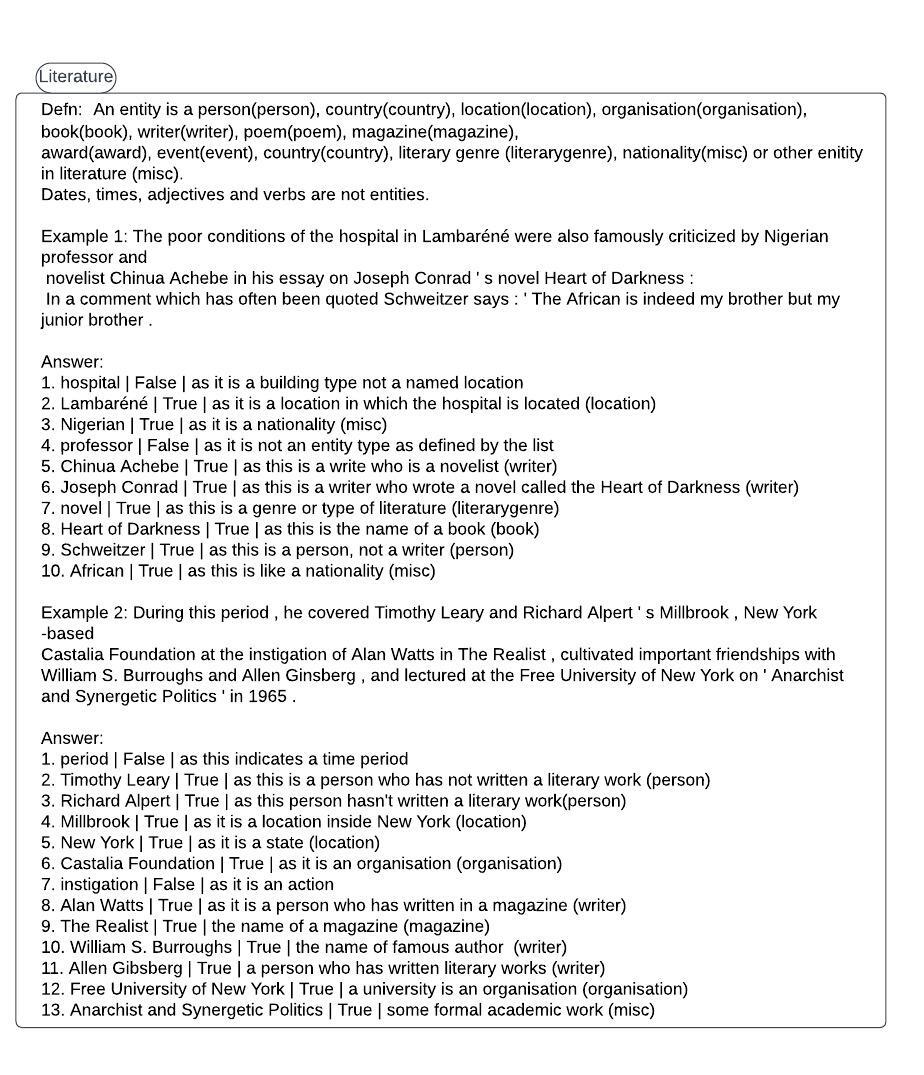}
    \caption{Literature Input Prompt}
\end{figure*}

\begin{figure*}
    \centering \includegraphics{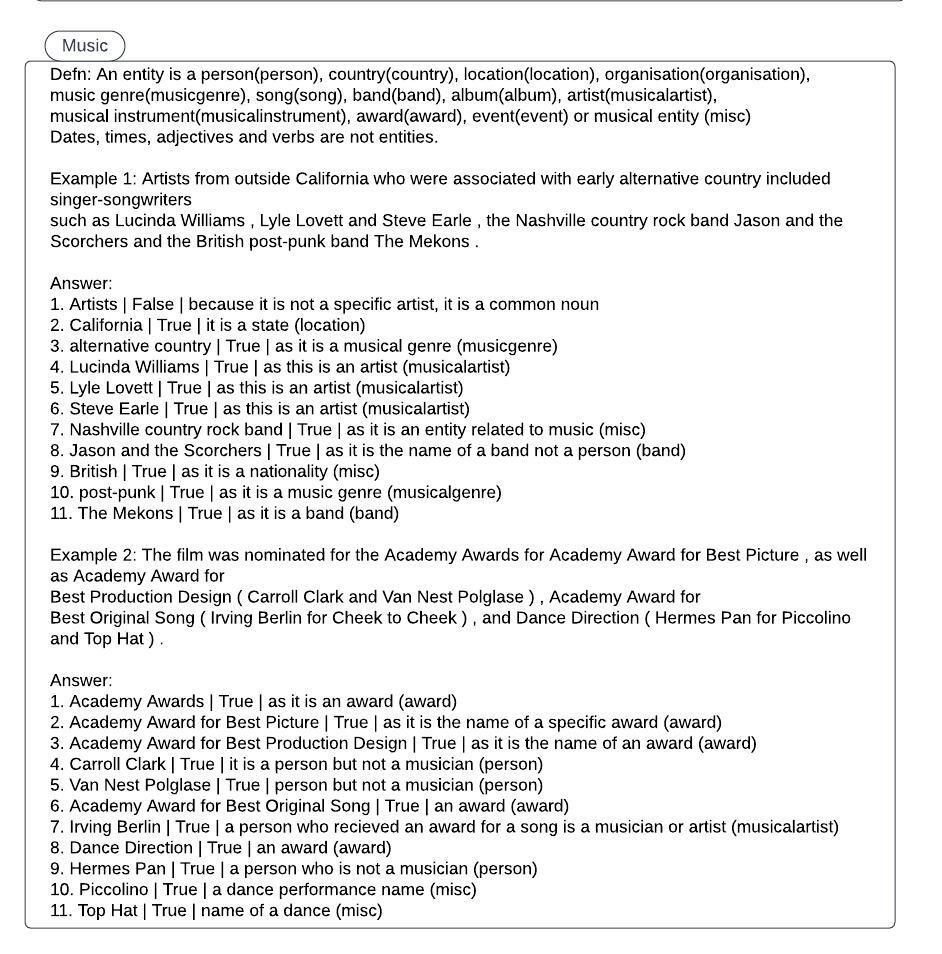}
    \caption{Music Input Prompt}
\end{figure*}

\begin{figure*}
    \centering \includegraphics{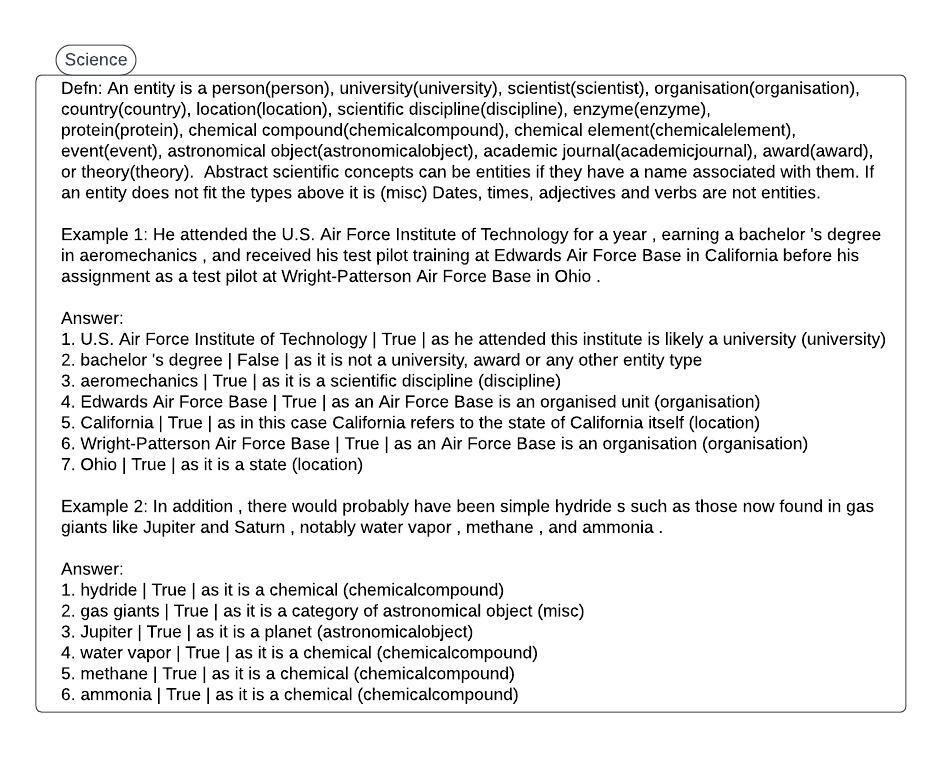}
    \caption{Science Input Prompt}
\end{figure*}

\begin{figure*}
    \centering \includegraphics{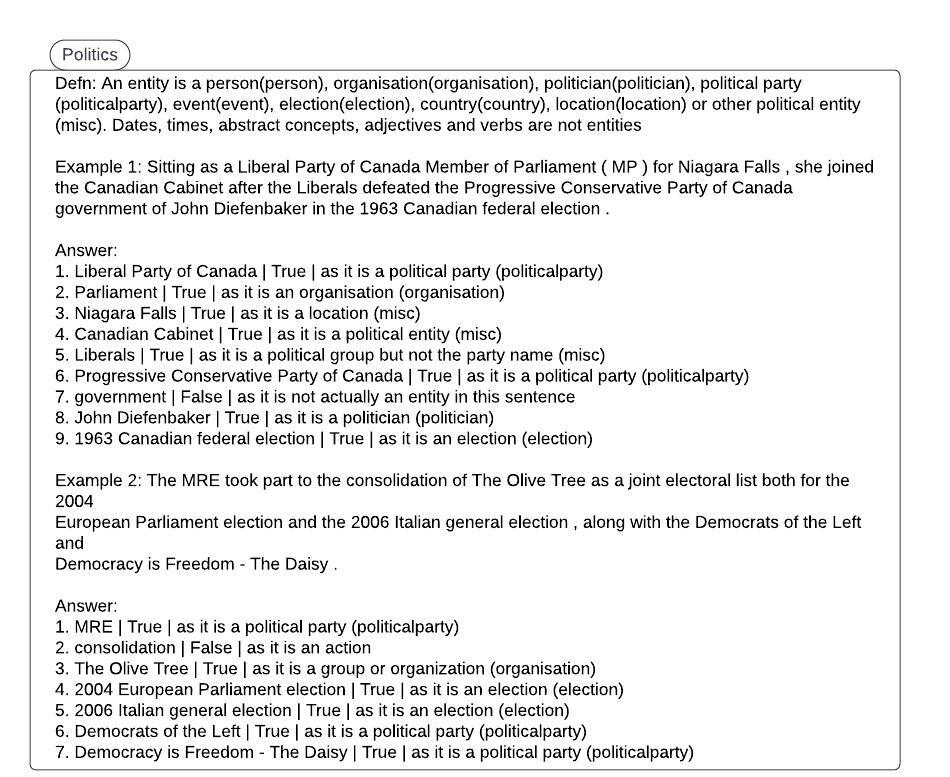}
    \caption{Politics Input Prompt}
\end{figure*}

\end{document}